\documentclass[10pt,onecolumn,twoside]{article}

\usepackage{amssymb} 
\usepackage{amsfonts}
\usepackage{amsmath} 
\usepackage{cite}
\usepackage{bm} 
\usepackage{graphicx} 
\usepackage{cite} 
\usepackage{authblk} 
\usepackage{algorithmic}
\usepackage{algorithm}
\usepackage{float}
\usepackage{fullpage}

\def\defn{\,\triangleq\,} 
\def\d{\, \mathrm{d}} 

\def\argmin{\mathop{\mathrm{arg\,min}}} 

\def\rbf{\mathbf{r}}

\def\xbf{\mathbf{x}}
\def\ybf{\mathbf{y}}
\def\cbf{\mathbf{c}}
\def\ebf{\mathbf{e}}
\def\bbf{\mathbf{b}}
\def\gbf{\mathbf{g}}
\def\zbf{\mathbf{z}}
\def\Hbf{\mathbf{H}}
\def\Sbf{\mathbf{S}}

\def\Psibf{\mathbf{\Psi}}
\def\Ibf{\mathbf{I}}

\def\xbfhat{\widehat{\mathbf{x}}}
\def\cbfhat{\widehat{\mathbf{c}}}

\def\Ecal{\mathcal{E}}

\def\Ccal{\mathcal{C}}

\def\Tcal{\mathcal{T}}

\def\Ncal{\mathcal{N}}

\def\Rcal{\mathcal{R}}

\def\R{\mathbb{R}}

\def\Z{\mathbb{Z}}

\def\prox{\mathrm{prox}}
\def\proj{\mathrm{proj}}
\def\diag{\mathrm{diag}}
\def\T{\mathrm{T}}
\def\half{{\textstyle\frac{1}{2}}}

\begin{document}


\title{Learning optimal nonlinearities for iterative thresholding algorithms}


\author{Ulugbek~S.~Kamilov%
\thanks{U.~S.~Kamilov (email: kamilov@merl.com) and H.~Mansour (email: mansour@merl.com)
are with Mitsubishi Electric Research Laboratories, 201 Broadway, Cambridge,
MA 02140, USA.}
\hspace{0.1em} and Hassan Mansour}

\maketitle 

\begin{abstract}
Iterative shrinkage/thresholding algorithm (ISTA) is a well-studied method for finding sparse solutions to ill-posed inverse problems. In this letter, we present a data-driven scheme for learning optimal thresholding functions for ISTA. The proposed scheme is obtained by relating iterations of ISTA to layers of a simple deep neural network (DNN) and developing a corresponding error backpropagation algorithm that allows to fine-tune the thresholding functions. Simulations on sparse statistical signals illustrate potential gains in estimation quality due to the proposed data adaptive ISTA.
\end{abstract}


\section{Introduction}
\label{Sec:Introduction}

The problem of estimating an unknown signal from noisy linear observations is fundamental in signal processing. The estimation task is often formulated as the linear inverse problem
\begin{equation}
\label{Eq:LinearModel}
\ybf = \Hbf \xbf + \ebf,
\end{equation}
where the objective is to recover the unknown signal $\xbf \in \R^N$ from the noisy measurements $\ybf \in \R^M$. The matrix $\Hbf \in \R^{M \times N}$ models the response of the acquisition device and the vector $\ebf \in \R^M$ represents the measurement noise, which is often assumed to be independent and identically distributed (i.i.d.) Gaussian. 

A standard approach for solving ill-posed linear inverse problems is the regularized least-squares estimator
\begin{equation}
\label{Eq:RegularizedLS}
\xbfhat = \argmin_{\xbf \in \R^N} \left\{ \frac{1}{2}\|\ybf - \Hbf\xbf\|_{\ell_2}^2 + \lambda \Rcal(\xbf) \right\},
\end{equation}
where $\Rcal$ is a regularizer that promotes solutions with desirable properties and $\lambda > 0$ is a parameter that controls the strength of regularization. In particular, sparsity-promoting regularization, such as $\ell_1$-norm penalty ${\Rcal(\xbf) \defn \|\xbf\|_{\ell_1}}$, has proved to be successful in a wide range of applications where signals are naturally sparse. Regularization with the $\ell_1$-norm is an essential component of compressive sensing theory~\cite{Candes.etal2006, Donoho2006}, which establishes conditions for accurate estimation of the signal from $M < N$ measurements.

The minimization~\eqref{Eq:RegularizedLS} with sparsity promoting penalty is a non-trivial optimization task. The challenging aspects are the non-smooth nature of the regularization term and the massive quantity of data that typically needs to be processed. Proximal gradient methods~\cite{Bauschke.Combettes2010} such as iterative shrinkage/thresholding algorithm (ISTA)~\cite{Figueiredo.Nowak2003, Bect.etal2004, Daubechies.etal2004} or alternativng direction method of multipliers (ADMM)~\cite{Eckstein.Bertsekas1992, Boyd.etal2011} are standard approaches to circumvent the non-smoothness of the regularizer while simplifying the optimization problem into a sequence of computationally efficient operations.

For the problem~\eqref{Eq:RegularizedLS}, ISTA can be written as
\begin{subequations}
\label{Eq:ISTA}
\begin{align}
\label{Eq:ISTA1}
\zbf^t &\leftarrow (\Ibf - \gamma \Hbf^\T\Hbf) \xbf^{t-1} + \gamma \Hbf^\T \ybf \\
\label{Eq:ISTA2}
\xbf^{t} &\leftarrow \Tcal(\zbf^t; \gamma \lambda),
\end{align}
\end{subequations}
where $\Ibf$ is the identity matrix and $\gamma > 0$ is a step-size that can be set to $\gamma =1/L$ with $L \defn \lambda_{\text{\tiny max}} (\Hbf^\T \Hbf)$ to ensure convergence~\cite{Beck.Teboulle2009}. Iteration~\eqref{Eq:ISTA} combines the gradient descent step~\eqref{Eq:ISTA1} with a proximal operator~\eqref{Eq:ISTA2} that reduces to a pointwise nonlinearity
\begin{subequations}
\begin{align}
\Tcal(z; \lambda) 
&= \prox_{\gamma \Rcal}(z) \\
&\defn \argmin_{x \in \R} \left\{\frac{1}{2}(x - z)^2 + \lambda \Rcal(x)\right\}.
\end{align}
\end{subequations}
for convex and separable regularizers such as the $\ell_1$-norm penalty.

\begin{figure}[t]
\begin{center}
\includegraphics[width=0.55\linewidth]{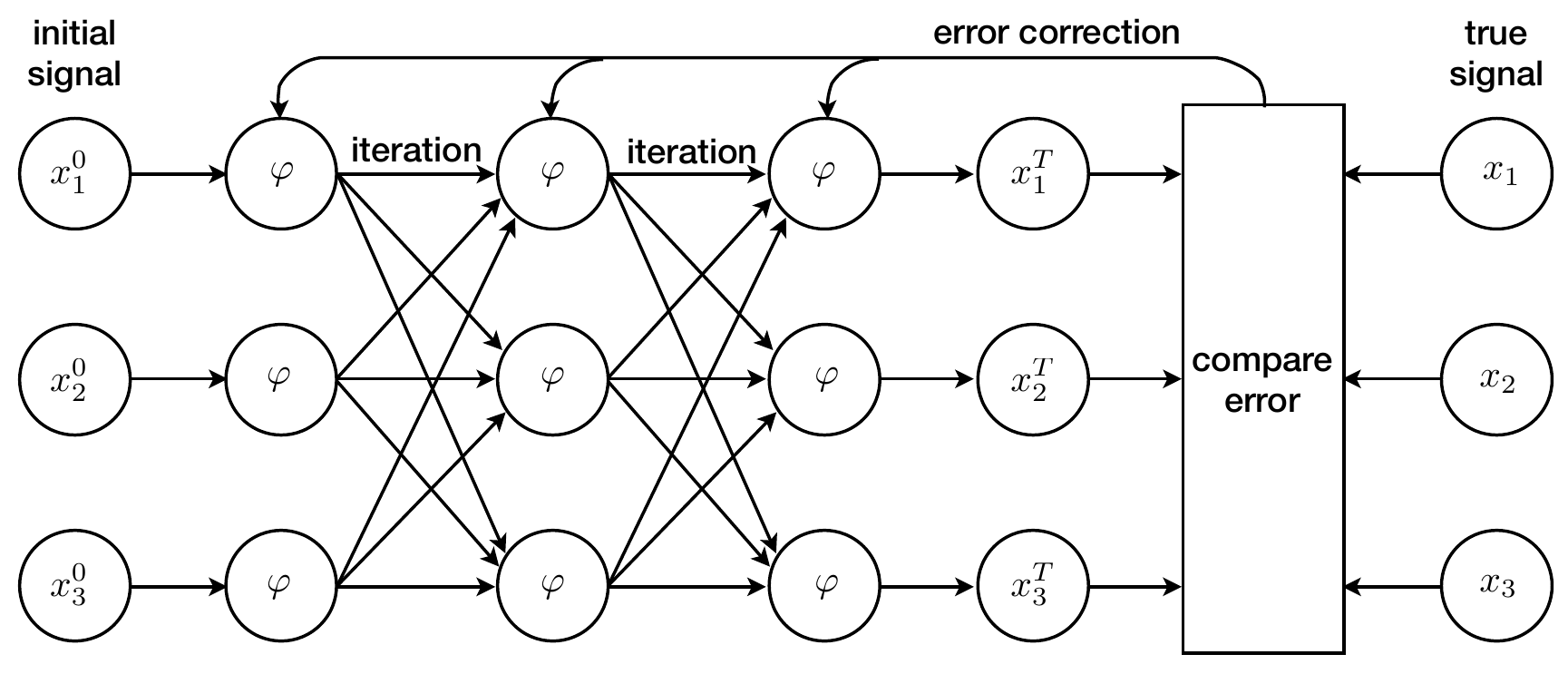}
\end{center}
\caption{Visual representation of the optimization scenario considered in this letter. ISTA with a pointwise nonlinearity $\varphi$ is initialized with a signal $\xbf^0$ which results in the estimate $\xbf^T$ after $T$ iterations. The algorithm proposed here allows to efficiently refine $\varphi$ by comparing $\xbf^T$ against the true signal $\xbf$ from a set of training examples.}
\label{Fig:DeepNN}
\end{figure}

In this letter, we consider the problem of learning an optimal nonlinearity $\Tcal$ for ISTA given a set of $L$ training examples $\{\xbf_\ell, \ybf_\ell\}_{\ell \in [1, \dots, L]}$. Specifically, as illustrated in Fig.~\ref{Fig:DeepNN}, we interpret iteration~\eqref{Eq:ISTA} as a simple deep neural network (DNN)~\cite{Bishop1995, LeCun.etal2015} with $T$ layers and develop an efficient algorithm that allows to determine optimal $\Tcal$ directly from data. Simulations on sparse statistical signals show that data adaptive ISTA substantially improves over the $\ell_1$-regularized reconstruction by approaching the performance of the minimum mean squared error (MMSE) estimator. 


\section{Related Work}
\label{Sec:RelatedWork}

\section{Related Work}
\label{Sec:RelatedWork}

Starting from the early works~\cite{Figueiredo.Nowak2003, Bect.etal2004, Daubechies.etal2004}, iterative thresholding algorithms have received significant attention in the context of sparse signal estimation. Accelerated variants of ISTA were proposed by, among others, Bioucas-Dias and Figueiredo~\cite{Bioucas-Dias.Figueiredo2007}, and Beck and Teboulle~\cite{Beck.Teboulle2009}. The method has also inspired approximate message passing (AMP) algorithm by Donoho \textit{et al.}~\cite{Donoho.etal2009}, as well as its Bayesian extensions~\cite{Donoho.etal2010, Rangan2011}. In particular, it was shown that, in the compressive sensing setting, one can obtain an optimal estimation quality by adapting the thresholding function of AMP to the statistics of the signal~\cite{Kamilov2015}. The primary difference of the work here is that the optimal thresholding functions are learned directly from independent realizations of the data, rather than being explicitly designed to the assumed statistics. The other difference is that ISTA, unlike AMP, contains no inherent assumptions on the randomness of the measurement matrix $\Hbf$~\cite{Rangan.etal2015}.

More recently, several authors have considered relating iterative algorithms to DNNs. For example, in the context of sparse coding, Gregor and LeCun~\cite{Gregor.LeCun2010} proposed to accelerate ISTA by learning the matrix $\Hbf$ from data. The idea was further refined by Sprechmann \textit{et al.}~\cite{Sprechmann.etal2012} by considering an unsupervised learning approach and incorporating a structural sparsity model for the signal. In the context of the image deconvolution problem, Schmidt and Roth~\cite{Schmidt.Roth2014} proposed a scheme to jointly learn iteration dependent dictionaries and thresholds for ADMM. Similarly, Chen \textit{et al.}~\cite{Chen.etal2015} proposed to parametrize nonlinear diffusion models, which are related to the gradient descent method, and learned the parameters given a set of training images. This letter extends those works by specifically learning separable thresholding functions for ISTA. Unlike the matrices $\Hbf$, thresholding functions relate directly to the underlying statistical distributions of i.i.d. signals $\xbf$. Furthermore, by optimizing for the same nonlinearity across iterations, we obtain the MSE optimal ISTA for a specific statistical distribution of data, which, in turn, allows us to evaluate the best possible reconstruction achievable by ISTA.


\section{Main Results}
\label{Sec:MainResults}

By defining a matrix $\Sbf \defn \Ibf - \gamma \Hbf^\T\Hbf$, vector $\bbf \defn \gamma \Hbf^T\ybf$, as well as nonlinearity $\varphi(\cdot) \defn \Tcal(\cdot, \gamma\lambda)$, we can re-write ISTA as follows
\begin{subequations}
\label{Eq:DISTA}
\begin{align}
z^t_m &\leftarrow \sum_{n = 1}^N S_{m n}x^{t-1}_n + b_m \\
x_m^t &\leftarrow \varphi(z_m^t),
\end{align}
\end{subequations}
where $m \in [1, \dots, N]$.

\subsection{Problem Formulation}
\label{Sec:ProblemFormulation}

Our objective is now to design an efficient algorithm for adapting the function $\varphi$, given a set of $L$ training examples $\{\xbf_\ell, \ybf_\ell \}_{\ell \in [1, \dots, L]}$, as well as by assuming a fixed number iterations $T$. In order to devise a computational approach for tuning $\varphi$, we adopt the following parametric representation for the nonlinearities
\begin{equation}
\label{Eq:NonLinearity}
\varphi(z) \defn \sum_{k = -K}^K c_k \psi\left(\frac{z}{\Delta} - k\right),
\end{equation}
where $\cbf \defn \{c_k\}_{k \in [-K, \dots, K]}$, are the coefficients of the representation and $\psi$ is a basis function, to be discussed shortly, positioned on the grid $\Delta [-K, -K+1, \dots, K] \subseteq \Delta \Z$.
We can reformulate the learning process in terms of coefficients $\cbf$ as follows
\begin{equation}
\label{Eq:MinProblem}
\cbfhat  =\argmin_{\cbf \in \Ccal }\left\{\frac{1}{L}\sum_{\ell = 1}^L \Ecal(\cbf, \xbf_\ell, \ybf_\ell)\right\}
\end{equation}
where $\Ccal \subseteq \R^{2K+1}$ is is used to incorporate some prior constraints on the coefficients and $\Ecal$ is a cost functional that guides the learning. The cost functional that interests us in this letter is the MSE defined as
\begin{equation}
\label{Eq:MMSECost}
\Ecal(\cbf, \xbf, \ybf) \defn \frac{1}{2}\|\xbf - \xbf^T(\cbf, \ybf)\|_{\ell_2}^2,
\end{equation}
where $\xbf^T$ is the solution of ISTA at iteration $T$, which depends on both the coefficients $\cbf$ and the given data vector $\ybf$. Given a large number of independent and identically distributed realizations of the signals $\{\xbf_\ell, \ybf_\ell\}$, the empirical MSE is expected to approach the true MSE of ISTA for nonlinearities of type~\eqref{Eq:NonLinearity}. Thus, by solving the minimization problem~\eqref{Eq:MinProblem} with the cost~\eqref{Eq:MMSECost}, we are seeking the MMSE variant of ISTA for a given statistical distribution of the signal $\xbf$ and measurements $\ybf$.

\subsection{Optimization}
\label{Sec:Optimization}

For notational simplicity, we now consider the scenario of a single training example and thus drop the indices $\ell$ from the subsequent derivations. The generalization of the final formula to an arbitrary number of training samples $L$ is straightforward.

We would like to minimize the following cost
\begin{equation}
\Ecal(\cbf) \defn \frac{1}{2}\|\xbf - \xbf^T(\cbf)\|_{\ell_2}^2 = \frac{1}{2} \sum_{m = 1}^N (x_m - x_m^T(\cbf))^2,
\end{equation}
where we dropped the explicit dependence of $\xbf^T$ on $\ybf$ for notational convenience. The optimization of the coefficients is performed via the projected gradient iterations
\begin{equation}
\cbf^i = \proj_\Ccal(\cbf^{i-1} - \mu\nabla \Ecal(\cbf^{i-1})),
\end{equation}
where $i = 1, 2, 3, \dots,$ denotes the iteration number of the training process, $\mu > 0$ is the step-size, which is also called the learning rate, and $\proj_\Ccal$ is an orthogonal projection operator onto the convex set $\Ccal$. 


\begin{algorithm}[t]
\caption{\textbf{Backpropagation} for evaluating $\nabla \Ecal(\cbf)$}
\label{Algo:Backprop}
\textbf{input: } measurements $\ybf$, signal $\xbf$, current value of coefficients $\cbf$, and number of ISTA iterations $T$.\\
\textbf{output: } the gradient $\nabla \Ecal(\cbf)$. \\
\textbf{algorithm: }
\begin{enumerate}
\item Run $T$ iterations of ISTA in eq.~\eqref{Eq:DISTA} by storing intermediate variables $\{\zbf^t\}_{t \in [1, \dots, T]}$ and the final estimate $\xbf^T$.
\item \emph{Initialize: } Set $t = T$, $\rbf^T = \xbf^T-\xbf$, and $\gbf^T = 0$.
\item \emph{Compute: }
\begin{subequations}
\begin{align}
\gbf^{t-1} &\leftarrow \gbf^{t} + [\Psibf^t]^\T\rbf^t\\
\rbf^{t-1} &\leftarrow \Sbf^\T\diag(\varphi^\prime(\zbf^{t}))\rbf^t
\end{align}
\end{subequations}
\item If $t = 0$, return $\nabla \Ecal(\cbf) = \gbf^0$, otherwise, set $t \leftarrow t-1$ and proceed to step 3).
\end{enumerate}
\end{algorithm}

We now devise an efficient error backpropagation algorithm for computing the derivatives of $\Ecal$ with respect to coefficients $\cbf$. First, note that we can write the iteration~\eqref{Eq:DISTA} with the nonlinearity~\eqref{Eq:NonLinearity} as follows
\begin{equation}
\label{Eq:Step}
x_m^t = \varphi(z_m^t) = \sum_{k = -K}^K c_k \psi\left(\frac{z_m^t}{\Delta} - k\right),
\end{equation}
for all $m \in [1, \dots, N]$. The gradient can be obtained by evaluating
\begin{equation}
\label{Eq:Grad}
\nabla \Ecal(\cbf) = \left[\frac{\partial}{\partial \cbf} \xbf^T(\cbf)\right]^\T(\xbf^T(\cbf)-\xbf),
\end{equation}
where we define
\begin{subequations}
\label{Eq:DerivativeISTA}
\begin{align}
\frac{\partial}{\partial \cbf} \xbf^t(\cbf) 
&\defn \left[\frac{\partial \xbf^t}{\partial c_{-K}} \dots \frac{\partial \xbf^t}{\partial c_{K}}\right] \\
&= 
\begin{bmatrix}
\frac{\partial x_1^t}{\partial c_{-K}} & \dots & \frac{\partial x_1^t}{\partial c_{K}}\\[0.3em]
\vdots & \ddots & \vdots \\[0.3em]
\frac{\partial x_N^t}{\partial c_{-K}} & \dots & \frac{\partial x_N^t}{\partial c_{K}}
\end{bmatrix}.
\end{align}
\end{subequations}
By differentiating~\eqref{Eq:Step} and simplifying the resulting expression, we obtain
\begin{equation}
\frac{\partial x_m^t}{\partial c_k} = \Psi^t_{mk} + \varphi^\prime(z_m^t) \sum_{n = 1}^N S_{mn} \left[\frac{\partial x_n^{t-1}}{\partial c_k}\right],
\end{equation}
where we defined a matrix $\Psi^t_{m k} \defn \psi(z_m^t/\Delta - k)$.
Then, for any vector $\rbf \in \R^N$, we obtain
\begin{align}
\sum_{m = 1}^N \left[\frac{\partial x_m^t}{\partial c_k}\right] r_m &= \sum_{m = 1}^N \Psi_{mk}^t r_m \\
&+ \sum_{n = 1}^N \left[\frac{\partial x_n^{t-1}}{\partial c_k}\right] \sum_{m = 1}^N S_{mn} r_m \varphi^\prime(z_m^t), \nonumber
\end{align}
which translates to the following vector equation
\begin{equation}
\label{Eq:GradRecursion}
\left[\frac{\partial \xbf^t}{\partial \cbf}\right]^\T \rbf = [\Psibf^t]^\T \rbf + \left[\frac{\partial \xbf^{t-1}}{\partial \cbf}\right]^\T \Sbf^\T \diag(\varphi^\prime(\zbf^t))\rbf,
\end{equation}
where the operator $\diag(\gbf)$ creates a matrix and places the vector $\gbf$ into its main diagonal. Note that since the initial estimate $\xbf^0$ does not depend on $\cbf$, we have that
\begin{equation}
\label{Eq:Initialization}
\frac{\partial \xbf^0}{\partial \cbf} = 0.
\end{equation}
By applying the equation~\eqref{Eq:GradRecursion} recursively starting from $t = T$ and by using~\eqref{Eq:Initialization}, we obtain
\begin{align*}
\left[\frac{\partial \xbf^T}{\partial \cbf}\right]^\T \rbf^T 
&= \underbrace{[\Psibf^T]^\T \rbf^T}_{\defn \gbf^{T-1}} + \left[\frac{\partial \xbf^{T-1}}{\partial \cbf}\right]^\T \underbrace{\Sbf^\T \diag(\varphi^\prime(\zbf^T))\rbf^T}_{\defn \rbf^{T-1}} \\
&= \gbf^{T-1} + \left[\frac{\partial \xbf^{T-1}}{\partial \cbf}\right]^\T \rbf^{T-1} \\
&= \underbrace{\gbf^{T-1} + [\Psibf^{T-1}]^\T\rbf^{T-1}}_{\defn \gbf^{T-2}} \\
&\quad+ \left[\frac{\partial \xbf^{T-2}}{\partial \cbf}\right]^\T \underbrace{\Sbf^\T \diag(\varphi^\prime(\zbf^{T-1}))\rbf^{T-1}}_{\defn \rbf^{T-2}} \\
&= \gbf^{T-2} + \left[\frac{\partial \xbf^{T-2}}{\partial \cbf}\right]^\T \rbf^{T-2} = \dots \\
&= \gbf^0 + \left[\frac{\partial \xbf^0}{\partial \cbf}\right]^\T \rbf^0 = \gbf^0.
\end{align*}
This suggests the error backpropagation algorithm summarized in Algorithm~\ref{Algo:Backprop} that allows one to obtain~\eqref{Eq:Grad}.


\begin{algorithm}[t]
\caption{\textbf{Online learning} for solving \eqref{Eq:MinProblem}}
\label{Algo:SGD}
\textbf{input: } set of $L$ training examples $\{\xbf_\ell, \ybf_\ell\}_{\ell \in [1,\dots,L]}$, learning rate $\mu > 0$, and constraint set $\Ccal \subseteq \R^{2K+1}$.\\
\textbf{output: } nonlinearity $\varphi$ specified by learned coefficients $\cbfhat$. \\
\textbf{algorithm: }
\begin{enumerate}
\item \emph{Initialize: } Set $i = 1$ and select $\cbf^0 \in \Ccal$.
\item Select a small subset $\{\xbf_\ell\}$ and $\{\ybf_\ell\}$, uniformly at random, from the set of $L$ training examples.
\item Using the selected training examples, update $\varphi$ as follows
\begin{equation}
\cbf^i \leftarrow \proj_\Ccal (\cbf^{i-1} - \mu \nabla \Ecal(\cbf^{i-1}))
\end{equation}
\item Return $\cbfhat = \cbf^i$ if a stopping criterion is met, otherwise set $i \leftarrow i+1$ and proceed to step 2).
\end{enumerate}
\end{algorithm}

The remarkable feature of Algorithm~\ref{Algo:Backprop} is that it allows one to efficiently evaluate the gradient of ISTA with respect to the nonlinearity $\varphi$. Its computational complexity is equivalent to running a single instance of ISTA, which is a first-order method, known to be scalable to very large scale inverse problems. Finally, equipped with Algorithm~\ref{Algo:Backprop}, nonlinearity $\varphi$ can easily be optimized by using an online learning approach~\cite{Zinkevich2003} summarized in Algorithm~\ref{Algo:SGD}.

\subsection{Representation with B-Splines}
\label{Sec:Splines}

In our implementation, we represent the nonlinearity $\varphi$ in terms of its expansion with polynomial B-Splines (For more details, see an extensive review of B-Spline interpolation by Unser~\cite{Unser1999}). Accordingly, our basis function corresponds to $\psi = \beta^d$, where $\beta^d$ refers to a B-Spline of degree $d \geq 0$. Within the family of polynomial splines, cubic B-Splines 
\begin{equation}
\beta^3(z) =
\begin{cases}
\frac{2}{3}-|z|^2 + \frac{|z|^3}{2}  & \text{when } 0 \leq |z| \leq 1\\
\frac{1}{6}(2 - |z|)^3 & \text{when } 1 \leq |z| \leq 2 \\
0 & \text{when } 2 \leq |z|,
\end{cases}
\end{equation}
tend to be the most popular in applications---perhaps due to their minimum curvature property~\cite{Unser1999}. B-Splines are very easy to manipulate. For instance, their derivatives are computed through the following formula
\begin{equation}
\frac{\d}{\d z} \beta^d(z) = \beta^{d-1}(z+\half)-\beta^{d-1}(z-\half),
\end{equation}
which simply reduces the degree by one. By applying this formula to the expansion of $\varphi$, we can easily obtain a closed form expression for $\varphi^\prime$ in terms of quadratic B-Splines
\begin{equation}
\beta^2(z) =
\begin{cases}
\frac{3}{4}-|z|^2  & \text{when } 0 \leq |z| \leq \frac{1}{2}\\
\frac{9}{8}-\frac{1}{2}|z|(3-|z|) & \text{when } \frac{1}{2} \leq |z| \leq \frac{3}{2} \\
0 & \text{when } \frac{3}{2} \leq |z|.
\end{cases}
\end{equation}


\section{Experiments}
\label{Sec:Experiments}

\begin{figure}[t]
\begin{center}
\includegraphics[width=0.55\linewidth]{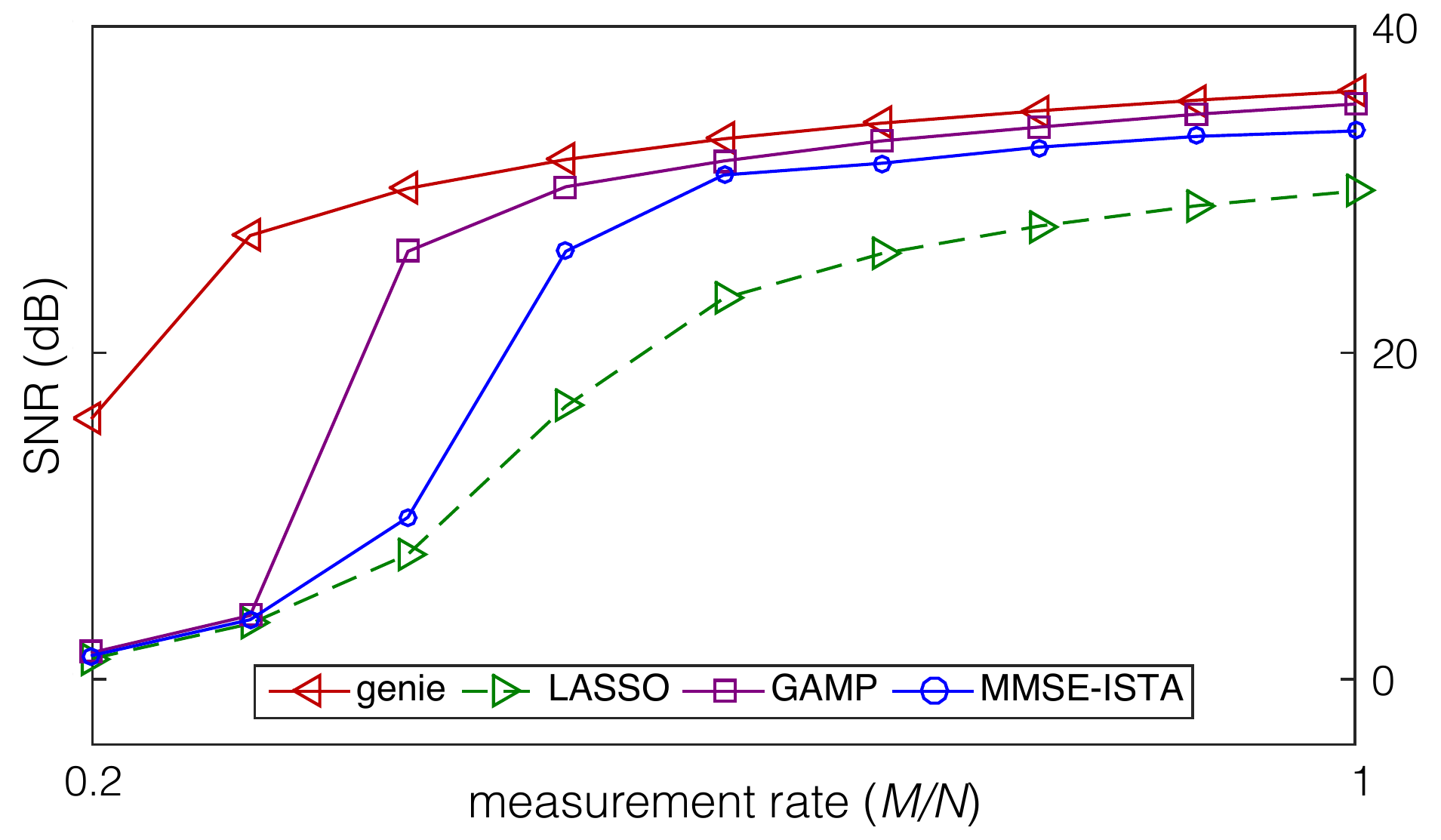}
\end{center}
\caption{Quantitative evaluation on sparse signals. Average SNR is plotted against the measurement rate $M/N$ when recovering $N = 512$ Benoulli-Gaussian signal $\xbf$ from measurements $\ybf$ under i.i.d. $\Hbf$. Note the proximity of MMSE-ISTA to the support-aware genie.}
\label{Fig:SNRs}
\end{figure}

\begin{figure}[t]
\begin{center}
\includegraphics[width=0.70\linewidth]{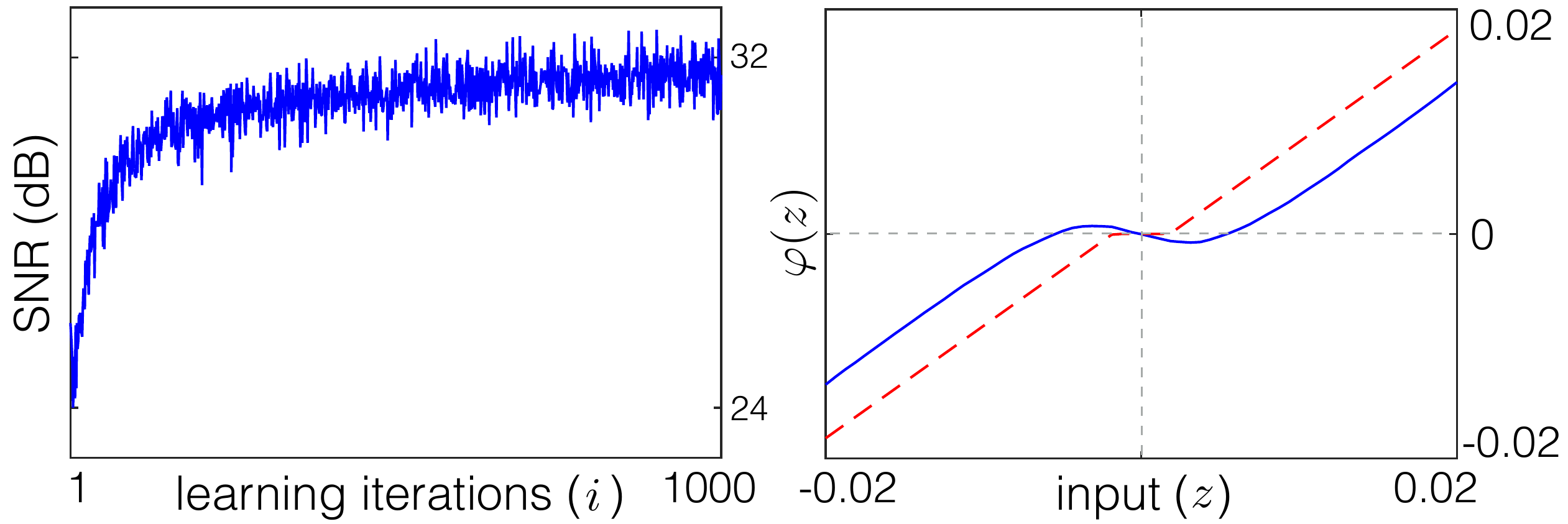}
\end{center}
\caption{Illustration of the learning process for $M/N = 0.7$. On the left side, SNR of training is plotted for each training iteration. On the right side, the final learned shrinkage (solid) is compared to the standard soft-thresholding under optimal $\lambda$.}
\label{Fig:Learning}
\end{figure}

To verify our learning scheme, we report results of ISTA with learned MSE optimal nonlinearities (denoted \textit{MMSE-ISTA}) on the compressive sensing recovery problem. In particular, we consider the estimation of a sparse Bernoulli-Gaussian signal $\xbf$ with an i.i.d. prior ${p_x(x_n) = \rho \Ncal(x_n, 0, 1) + (1-\rho) \delta(x_n)}$, where $\rho \in (0, 1]$ is the sparsity ratio, $\Ncal(\cdot, \mu, \sigma^2)$ is the Gaussian probability distribution function of mean $\mu$ and variance $\sigma^2$, and $\delta$ is the Dirac delta distribution. In our experiments, we fix the parameters to $N = 512$ and $\rho = 0.2$, and we numerically compare the signal-to-noise ratio (SNR) defined as ${\textrm{SNR (dB)} \defn 10\log_{10}\left(\|\xbf\|^2_{\ell_2}/\|\xbf - \xbfhat\|_{\ell_2}^2\right)}$,
for the estimation of $\xbf$ from linear measurements of form~\eqref{Eq:LinearModel}, where $\ebf$ has variance set to achieve SNR of $30$ dB, and where the measurement matrix $\Hbf$ is drawn with i.i.d. $\Ncal(0, 1/M)$ entries.

We compare results of MMSE-ISTA against three alternative methods. As the first reference method, we consider standard least absolute shrinkage and selection operator (\emph{LASSO})~\cite{Tibshirani1996} estimator, which corresponds to solving~\eqref{Eq:RegularizedLS} with $\ell_1$-norm regularizer. In addition to LASSO, we consider MMSE variant of the generalized AMP (\emph{GAMP}) algorithm~\cite{Rangan2011}, which is known to be nearly optimal for recovery of sparse signals from random measurements. Finally, we consider a support-aware MMSE estimator (\emph{genie}), which provides an upper bound on the reconstruction performance of any algorithm.

The regularization parameter $\lambda$ of LASSO was optimized for the best SNR performance. Similarly, the parameters of GAMP were set to their statistically optimal values. The implementation of LASSO is based on FISTA~\cite{Beck.Teboulle2009}. Both FISTA and GAMP were run for a maximum of $1000$ iterations or until convergence that was measured using the relative change in the solution in two successive iterations ${\|\xbf^t - \xbf^{t-1}\|_{\ell_2}/\|\xbf^{t-1}\|_{\ell_2} \leq 10^{-4}}$. The number of layers of MMSE-ISTA was set to $T = 200$. Learning was performed by using online learning in Algorithm~\ref{Algo:SGD} that was run for $1000$ iterations with the learning rate of $\mu = 10^{-4}$. The nonlinearity $\varphi$ was defined with $8000$ basis functions that were spread uniformly over the dynamic range of the signal and was initialized to correspond to the soft-thresholding function with optimal $\lambda$.

Figure~\ref{Fig:SNRs} reports the SNR performance of all algorithms under test after averaging the results of $1000$ Monte Carlo trials. The results show that the quality of estimated signal can be considerably boosted by using nonlinearities $\varphi$ that are adapted to the data. In particular, the SNR performance of MMSE-ISTA is significantly better than that of LASSO and is about 1 dB away from the SNR obtained by GAMP at higher values of $M/N$. Figure~\ref{Fig:Learning} illustrates the per-iteration evolution of SNR evaluated on the training sample during the learning process (left), as well as the final shape of the learned nonlinearity (right). As can be appreciated from these plots, the learning procedure deviates the shape of nonlinearity $\varphi$ from the soft-thresholding function, which leads to a significant increase in SNR of the solution.


\section{Conclusion}
\label{Sec:Conclusion}

The scheme developed in this letters is useful for optimizing the nonlinearities of ISTA given a set of independent realizations of data samples. By using this scheme, we were able to benchmark the best possible reconstruction achievable by ISTA for i.i.d. sparse signals. Specifically, in the context of compressive sensing, we showed that by optimizing the nonlinearities the performance of ISTA improves by several dBs and approaches that of the optimal estimator. Future  investigations of ISTA under optimal nonlinearities may lead to an improved understanding of the relationship between statistical estimators and iterative reconstruction algorithms.


\bibliographystyle{IEEEtran}


\end{document}